# Signature Recognition using Multi Scale Fourier Descriptor And Wavelet Transform

Ismail A. Ismail [1] , Mohammed A. Ramadan [2] , Talaat S. El danaf [3] and Ahmed H. Samak[4]

[1] Professor, Dean, College of Computers and Informatics ,Misr International University, , Egypt
[2] Professor, Department of Mathematics, Faculty of Science, Menofia University , Egypt
[3] Ass. Professor , Department of Mathematics, Faculty of Science, Menofia University , Egypt
[4] Ass. Lecturer , Department of Mathematics, Faculty of Science, Menofia University, Egypt
Email : eng_ahmed_Samak@yahoo.co.uk

**Abstract** This paper present a novel off-line signature recognition method based on multi scale Fourier Descriptor and wavelet transform . The main steps of constructing a signature recognition system are discussed and experiments on real data sets show that the average error rate can reach 1%. Finally we compare 8 distance measures between feature vectors with respect to the recognition performance.

a) Key words signature recognition , Fourier Descriptor , Wavelet transform , personal verification

## 1- INTRODUCTION

In the past decades, biometrics research has always been the focus of interests for scientists and engineers. It is an art of science to use physical and behavioral characteristics to verify or identify a person. Particularly, handwriting is believed to be singular, exclusive, personal for individuals. Handwriting signature is the most popular identification method socially and legally which has been used widely in the bank check and credit card transactions, document certification, etc.

The objective of signature recognition is to recognize the signer for the purpose of recognition or verification. Recognition is finding the identification of the signature owner. Verification is the decision about whether the signature is genuine or forgery as in figure 1 .

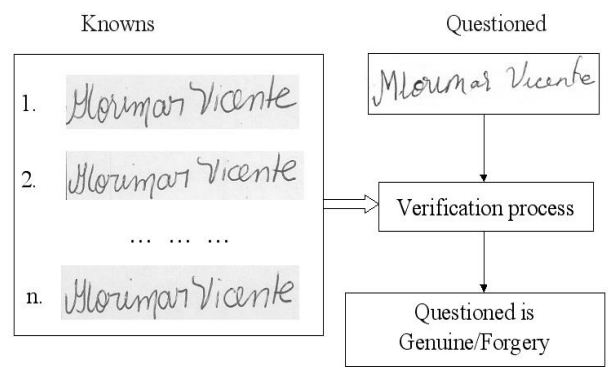

Figure 1: Verification task

Handwritten signature recognition can be divided into on-line (or dynamic) and off-line (or static) recognition. On-line recognition refers to a process that the signer uses a special pen called a stylus to create his or her signature, producing the pen locations, speeds and pressures, while off-line recognition just deals with signature images acquired by a scanner or a digital camera. In general, off-line signature recognition is a challenging problem. Off- line handwriting recognition systems are more difficult than online systems as dynamic information like duration, time ordering, number of strokes, and direction of writing are lost. But, offline systems have a significant advantage in that they don't require access to special processing devices when the signature is produced. This chapter deals with an off-line signature recognition and verification system.

In the last few decades, many approaches have been developed in the pattern recognition area, which approached the offline signature verification problem. Justino, Bortolozzi and Sabourin proposed an off-line signature verification system using Hidden Markov Model [1]. Zhang, Fu and Yan (1998) proposed handwritten signature verification system based on Neural 'Gas' based Vector Quantization [2]. Vélez, Sánchez and Moreno proposed robust off-line signature verification system using compression networks and positional cuttings [3].

Arif and Vincent (2003) concerned data fusion and its methods for an off-line signature verification problem which are Dempster-Shafer evidence theory, Possibility theory and Borda count method [4]. Chalechale and Mertins used line segment distribution of sketches for Persian signature recognition [5]. Sansone and Vento (2000) increased performance of signature verification system by a serial three stage multi-expert system [6].

Inan Güler and Majid Meghdadi ( 2008) proposed a method for the automatic handwritten signature verification (AHSV)is described. This method relies on global features that summarize different aspects of signature shape and dynamics of signature production. For designing the algorithm, they have tried to detect the signature without paying any attention to the thickness and size of it [7].

Jing Wen,BinFang, Y.Y.Tang and TaiPing Zhang (2009) presents two models utilizing rotation invariant structure features to tackle the problem. In principle, the elaborately extracted ring-peripheral features are able to describe internal and external structure changes of signatures periodically. In order to evaluate match score quantitatively, discrete fast Fourier transform is employed to eliminate phase shift and verification is conducted based on a distance model. In addition, the ring-hidden Markov model (HMM) is constructed to directly evaluate similar between test signature and training samples [8].

2- DATABASE

The signature database consists of 840 signature images, scanned at a resolution of 300 dpi,8-bit gray-scale. They are organized into 18 sets, and each set corresponds to one signature enrollment. There are 24 genuine and 24 forgery signatures in a set. Each volunteer was asked to sign his or her own signatures on a white paper 24 times. After this process had been done, we invited some people who are good at imitating other's handwritings. An examples of the database image are shown in figure 2.

3- PREPROCESSING

Any image-processing application suffers from noise like touching line segments, isolated pixels and smeared images. This noise may cause severe distortions in the digital image and hence ambiguous features and a correspondingly poor recognition and verification rate. Therefore, a preprocessor is used to remove noise. Preprocessing techniques eliminate much of the variability of signature data. Preprocessor also achieve the scaling and rotation invariant using slant normalization.

3-1 **Noise Reduction**

Standard noise reduction and isolated peak noise removal techniques, such as median-filtering [1], are used to clean the initial image. The median filter is a sliding-window spatial filter, it replaces the center value in the window with the median of all the pixel values in the window. The kernel is usually square but can be any shape. Figure 3 present example of noise reduction

3-2 **Slant Normalization**

Normalization is necessary to achieve the scaling and rotation invariant of the target images before the recognition phase. The scale normalization can be made by scaling the image along the *x* coordinate and *y* coordinate respectively to the prefixed size. For the slant normalization a moment based algorithm is described in [10]. The basic idea is to compute the major orientation or slant angle of the handwriting strokes according to second moments of foreground pixels and rotate the foreground pixels by the computed angle along the opposite direction such that the major orientation is horizontal. Figure 4 present examples of slant normalization

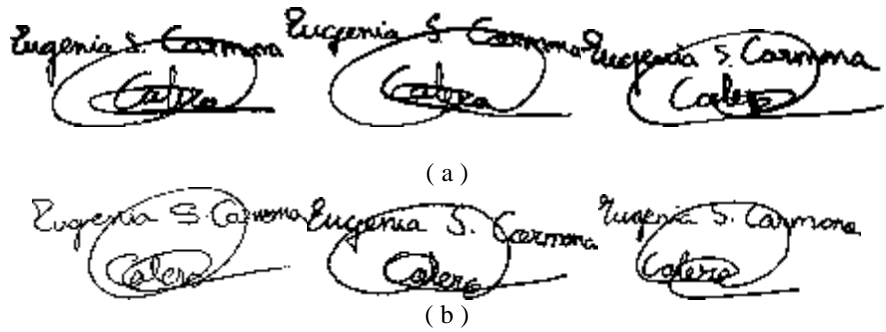

( a )

( b )

Figure 2 Database examples : ( a ) Genuine and ( b ) forgery signatures.

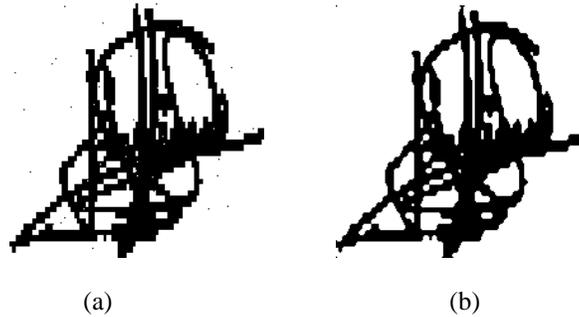

(a)                 (b)

Figure 3 : Noise reduction (a) noised image (b) filtered image

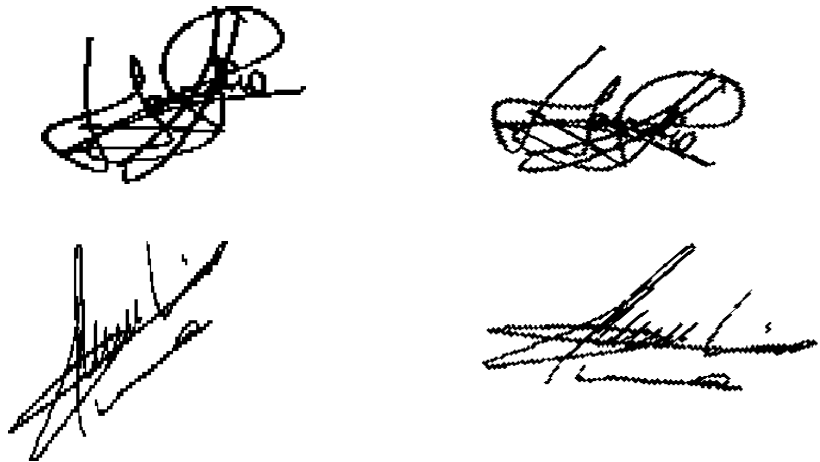

**Figure 4  Slant Normalization**

4- FEATURES EXTRACTION

The choice of a powerful set of features is crucial in signature identification systems. In our system, we use implement a new Multi scale Fourier descriptor using wavelet transform as discus in the next section

4-1 Multi scale Fourier descriptor using wavelet transform

The multi scale representation of the signature image can be achieved using wavelet transform. In discrete wavelet transform (DWT), the wavelet coefficient of the image $f(x, y)$ defined as

$$W_\vartheta(j_0,m,n) = \frac{1}{\sqrt{MN}} \sum_{x=0}^{M-1}\sum_{y=0}^{N-1} f(x,y)\vartheta_{j_0,m,n}(x,y)$$

$$W_\psi^i(j,m,n) = \frac{1}{\sqrt{MN}} \sum_{x=0}^{M-1}\sum_{y=0}^{N-1} f(x,y)\psi_{j,m,n}^i(x,y) \quad h=\{H,V,D\}$$

Where $J_0$ is arbitrary starting scale

The problem with the coefficients obtained from the wavelet transform is the fact that they not rotation invariant. Also the dimensionality of the feature vector depends on the signature image size . Therefore, the coefficient vectors of different signatures cannot be directly matched in the image retrieval. The proposed solution for this problem is to apply the Fourier transform to the coefficients obtained from the wavelet transform. In this way, the multi scale signature representation can be transformed to the frequency domain, in which normalization and matching are straightforward operations. Hence the benefits of multi scale representation and Fourier representation can be combined. The Multi scale Fourier descriptor is formed by applying the discrete Fourier transform of Eq

$$a_n = \frac{1}{N}\sum_{t=0}^{N-1} u(t)\exp(-j2\Pi nt/N) \text{ where } u(t) = w$$

This results in a set of Fourier coefficients { $a_n$ }, which is a representation of the signature region . Since image generated through rotation, translation and scaling (called similarity transform of a image ) of a same image are similar images , a image representation should be invariant to these operations. The selection of different start point on the image boundary to derive $u(t)$ should not affect the representation. From Fourier theory, the general form for the Fourier coefficients of a contour generated by translation, rotation, scaling and change of start point is given by :

$$a_n = \exp(jnt)\times\exp(j\phi)\times c\times a_n^{(0)} \qquad n\neq 0$$

where $a_n^0$ and $a_n$ are the Fourier coefficients of the original image and the similarity transformed image , respectively; $\exp(jnt)$, $\exp(j\Phi)$ and $s$ are the terms due to change of starting point, rotation and scaling. Except the DC component ($a_0$), all the other coefficients are not affected by translation. Now considering the following expression

$$b_n = \frac{a_n}{a_0} = \frac{\exp(jnt)\times\exp(j\phi)\times c\times a_n^{(0)}}{\exp(jt)\times\exp(j\phi)\times c\times a_0^{(0)}}$$

$$= \frac{a_n^{(0)}}{a_0^0}\exp[j(n-1)t] = b_n^{(0)}\exp[j(n-1)t]$$

where $b_n$ and $b_n^0$ are normalized Fourier coefficients of the derived image and the original image , respectively. The normalized coefficient of the derived image $b_n$ and that of the original image $b_n^0$ have only difference of $\exp[j(n-1)t]$. If we ignore the phase information and only use magnitude of the coefficients, then $|b_n|$ and $|b_n^0|$ are the same. In other words, $|b_n|$ is invariant to translation, rotation, scaling and change of start point. The set of magnitudes of the normalized Fourier coefficients of the signature image { $|b_n|$, $0 < n \leq N$ } can now be used as signature image descriptors, denoted as $\{FD_n\, 0 < n \leq N\}$. the next section discuses some distance measured can be used .

5- DISTANCE MEASURES

Let *X, Y* be feature vectors of length n. Then we can calculate the following distances between these feature vectors

Minkowski distance ( $L_P$ matrices)

$$d(X,Y) = L_p(X,y) = \left(\sum_{i=1}^n |x_i - y_i|\right)^{1/p};$$

Manhattan distance ( $L_1$ matrices , city block distance )

$$d(X,Y) = L_{p=1}(X,y) = \sum_{i=1}^n |x_i - y_i|;$$

Euclidean distance ( $L_2$ matrices)

$$d(X,Y) = L_{p=2}(X,y) = \sqrt{\sum_{i=1}^2 (X_i - y_i)^2};$$

Angle – based distance
$$d(X,Y) = -\cos(X,Y)$$

$$\cos(X,Y) = \frac{\sum_{i=1}^n x_i y_i}{\sqrt{\sum_{i=1}^n x_i^2 \sum_{i=1}^n y_i^2}};$$

Correlation coefficient- based distance
$$d(X,Y) = -r(X,Y)$$

$$r(X,Y) = \frac{n\sum_{i=1}^n x_i y_i - \sum_{i=1}^n x_i \sum_{i=1}^n y_i}{\sqrt{(n\sum_{i=1}^n x_i^2 - (\sum_{i=1}^n x_i)^2)(n\sum_{i=1}^n y_i^2 - (\sum_{i=1}^n y_i)^2)}};$$

Modified Manhattan distance

$$d(X,Y) = \frac{\sum_{i=}^n |x_i - y_i|}{\sum_{i=1}^n |x_i|\sum_{i=1}^n |y_i|}$$

Modified SSE-based distance

$$d(X,Y) = \frac{\sum_{i=1}^n (x_i - y_i)^2}{\sum_{i=1}^n x_i^2 \sum_{i=1}^n y_i^2}$$

6- EXPERIMENTAL RESULTS

This section reports some experimental results obtained using our method . In the following experiments, a total of 840 signature images. The experimental platform is the Intel core 2 duo 1.83 GHZ processor, 1G RAM, Windows vista , and the software is Matlab 7.0.0.1. The recognition performance is evaluated using different distance measure and different wavelet families as present in Table1,table 2 and table 3 . in the first table we use only FD as a features to present signature images ,also Table 2 use only Wavelet Transform finally Table 3 we use the new Multi scale Fourier descriptor using wavelet transform.

| Distance measures | Correct recognition rate |
|---|---|
| Minkowski distance | 92.6% |
| Manhattan distance | 96.2% |
| Euclidean distance | 95.4% |
| Angle – based distance | 93.4% |
| Correlation coefficient-based distance | 92.2% |
| Modified Manhattan distance | 95.6% |
| Modified SSE-based distance | 88.8% |

Table 1 : Recognition Performance using Fourier descriptor coefficients.

| Wav family / Distance measures | haar | DB2 | DB8 | DB15 | Sym8 |
|---|---|---|---|---|---|
| Minkowski distance | 89.8% | 91.4% | 92.8% | 93.6% | 93% |
| Manhattan distance | 93% | 93.4% | 94.4% | 95.2% | 94.6 % |
| Euclidean distance | 92.6% | 93.4% | 94.8% | 94.4% | 94.2% |
| Angle – based distance | 93.2% | 94.2% | 94.2% | 94.8% | 94.2% |
| Correlation coefficient- based distance | 92.2% | 92.8% | 93.4% | 94.2% | 93.8% |
| Modified Manhattan distance | 92% | 93.8% | 94.2% | 94.6% | 94% |
| Modified SSE-based distance | 93.2% | 94.2% | 94,6% | 95.6% | 94.4 % |

Table 2 : Recognition Performance Using wavelet transform .

| Wav family<br>Distance measures | haar | DB2 | DB8 | DB15 | Sym8 |
|---|---|---|---|---|---|
| Minkowski distance | 97% | 97.6% | 97.2% | 97% | 96.6% |
| Manhattan distance | 98.8% | 98.4% | 98% | 98.6% | 99% |
| Euclidean distance | 98.2% | 98.4% | 98.2% | 98.2% | 98.2% |
| Angle – based distance | 96.6% | 97.8% | 97% | 97.6% | 97.8% |
| Correlation coefficient- based distance | 95.8% | 96.6% | 96.6% | 97.2% | 97.2% |
| Modified Manhattan distance | 96.2% | 98% | 97.4% | 97.8% | 97.6% |
| Modified SSE-based distance | 92.8% | 96.2% | 96.8% | 98% | 95.8% |

Table 3 : Recognition Performance Using Multi scale Fourier descriptor and wavelet transform

7- CONCLUSIONS

In this publication we devolve a method for signature recognition based on multi scale Fourier Descriptor using wavelet transform. Recognition experiments were performed using the database containing 840 signature image . Our method tested using different wavelet family and various distance Measures. The best recognition results were achieved using sym8 wavelet family and Manhattan distance .